\documentclass[10pt,twocolumn,letterpaper]{article}

\usepackage{cvpr}
\usepackage{times}
\usepackage{epsfig}
\usepackage{graphicx}
\usepackage{amsfonts}
\usepackage{amsmath}
\usepackage{amssymb}

\usepackage{nicefrac}       
\usepackage{microtype}      

\usepackage{multirow}
\usepackage{verbatim}
\usepackage{color}
\usepackage{float}
\usepackage{enumitem}
\usepackage{booktabs}
\usepackage{tabulary,multirow,overpic,xcolor}
\usepackage[caption=false]{subfig}

\usepackage[british,UKenglish,USenglish,english,american]{babel}

\newcolumntype{x}[1]{>{\centering\arraybackslash}p{#1pt}}

\newcommand{\app}{\raise.17ex\hbox{$\scriptstyle\sim$}}

\newcolumntype{x}[1]{>{\centering\arraybackslash}p{#1pt}}

\newlength\savewidth
\newcommand{\tablestyle}[2]{\setlength{\tabcolsep}{#1}\renewcommand{\arraystretch}{#2}\centering\footnotesize}

\definecolor{citecolor}{RGB}{34,139,34}
\usepackage[pagebackref=true,breaklinks=true,letterpaper=true,colorlinks,
  citecolor=citecolor,bookmarks=false]{hyperref}

\cvprfinalcopy 

\def\cvprPaperID{740} 

\begin{document}

\title{Zero-shot Recognition via Semantic Embeddings and Knowledge Graphs}

\author{Xiaolong Wang\thanks{Indicates equal contribution.} 
\quad     \quad  Yufei Ye\footnotemark[1]  \quad  \quad  Abhinav Gupta  \\
The Robotics Institute, Carnegie Mellon University \\
}

\maketitle

\newcommand{\softmaxpp}{ConSE\xspace}
\newcommand{\n}{n}
\newcommand{\nway}{$\n$-way\xspace}
\newcommand{\eqdef}{\stackrel{\rm def}{=}}

\begin{abstract}
We consider the problem of zero-shot recognition: learning a visual classifier for a category with zero training examples, just using the word embedding of the category and its relationship to other categories, which visual data are provided. The key to dealing with the unfamiliar or novel category is to transfer knowledge obtained from familiar classes to describe the unfamiliar class. In this paper, we build upon the recently introduced Graph Convolutional Network (GCN) and propose an approach that uses both semantic embeddings and the categorical relationships to predict the classifiers. Given a learned knowledge graph (KG), our approach takes as input semantic embeddings for each node (representing visual category). After a series of graph convolutions, we predict the visual classifier for each category. During training, the visual classifiers for a few categories are given to learn the GCN parameters. At test time, these filters are used to predict the visual classifiers of unseen categories. We show that our approach is robust to noise in the KG. More importantly, our approach provides significant improvement in performance compared to the current state-of-the-art results (from $2 \sim 3\%$ on some metrics to whopping $20\%$ on a few).
\end{abstract}

\section{Introduction}

\begin{figure}[h]
    \centering
    \includegraphics[width=1.01\linewidth]{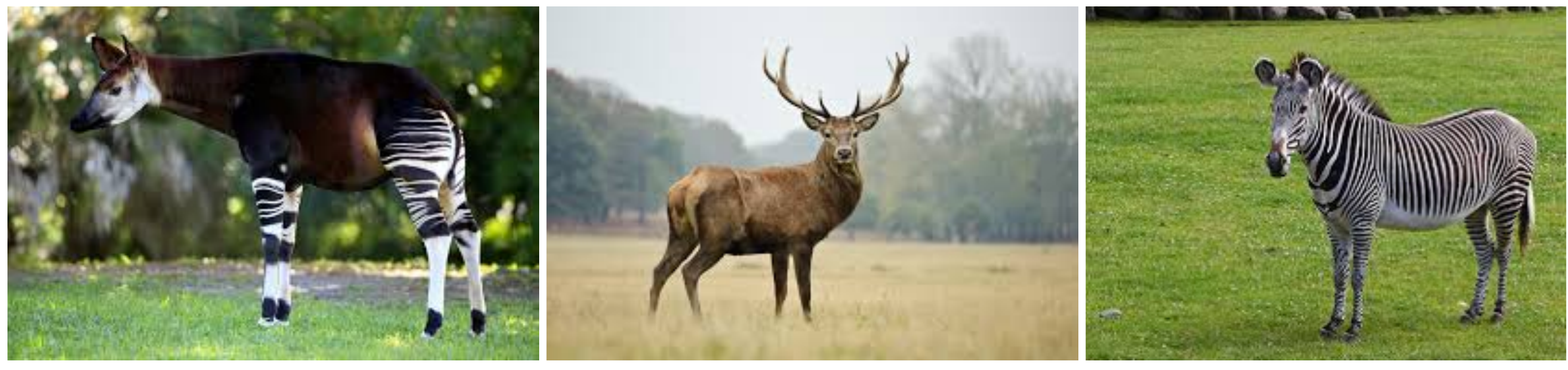}
    \caption{Can you find ``okapi'' in these images? Okapi is " zebra-striped four legged animal with a brown torso and a deer-like face". In this paper, we focus on the problem of zero-shot learning where visual classifiers are learned from semantic embeddings and relationships to other categories.}\label{fig:teaser}
\end{figure}

Consider the animal category ``okapi''. Even though we might have never heard of this category or seen visual examples in the past, we can still learn a good visual classifier based on the following description: "zebra-striped four  legged animal with a brown torso and a deer-like face" (Test yourself on figure~\ref{fig:teaser}). On the other hand, our current recognition algorithms still operate in closed world conditions: that is, they can only recognize the categories they are trained with. Adding a new category requires collecting  thousands of training examples and then retraining the classifiers. To tackle this problem, zero-shot learning is often used. 

The key to dealing with the unfamiliar or novel category is to transfer knowledge obtained from familiar classes to describe the unfamiliar classes (generalization). There are two paradigms of transferring knowledge. The first paradigm is to use implicit knowledge representations, i.e. semantic embeddings. In this approach, one learns a vector representation of different categories using text data and then learns a mapping between the vector representation to visual classifier directly~\cite{conse_iclr14,devise12}. However, these methods are limited by the generalization power of the semantic models and the mapping models themselves. It is also hard to learn semantic embeddings from structured information.

The alternative and less-explored paradigm for zero-shot learning is to use explicit knowledge bases or knowledge graphs. In this paradigm, one explicitly represents the knowledge as rules or relationships between objects. These relationships can then be used to learn zero-shot classifiers for new categories. The simplest example would be to learn visual classifiers of compositional categories. Given classifiers of primitive visual concepts as inputs, \cite{misra2017composing} applies a simple composition rule to generate classifiers for new complex concepts. However, in the general form, the relationships can be more complex than simple compositionality. An interesting question we want to explore is if we can use structured information and complex relationships to learn visual classifiers without seeing any examples.

In this paper, we propose to distill both the implicit knowledge representations (i.e. word embedding) and explicit relationships (i.e. knowledge graph) for learning visual classifiers of novel classes. We build a knowledge graph where each node corresponds to a semantic category. These nodes are linked via relationship edges. The input to each node of the graph is the vector representation (semantic embedding) of each category. We then use Graph Convolutional Network (GCN)~\cite{GCN17} to transfer information (message-passing) between different categories. Specifically, we train a 6-layer deep GCN that outputs the classifiers of different categories.

We focus on the task of image classification. We consider both of the test settings: (a) final test classes being only zero-shot classes (without training classes at test time); (b) at test time the labels can be either the seen or the unseen classes, namely ``generalized zero-shot setting'' ~\cite{Hariharan17,Chao16,Xian17}. We show surprisingly powerful results and huge improvements over classical baselines such as DeVise~\cite{devise12} , ConSE~\cite{conse_iclr14} ,and current state-of-the-art~\cite{Changpinyo17}. For example, on standard ImageNet with 2-hop setting, $43.7\%$ of the images retrieved by ~\cite{Changpinyo17} in top-10 are correct. Our approach retrieves $62.4\%$ images correctly. {\bf That is a whopping $18.7\%$ improvement over the current state-of-the-art.} More interestingly, we show that our approach scales amazingly well and giving a significant improvement as we increase the size of the knowledge graph even if the graph is noisy.

\section{Related Work}
\vspace{-0.05in}
With recent success of large-scale recognition systems~\cite{chensun}, the focus has now shifted to scaling these systems in terms of categories. As more realistic and practical settings are considered, the need for zero-shot recognition -- training visual classifiers without any examples -- has increased. Specifically, the problem of mapping text to visual classifiers is very interesting. 

Early work on zero-shot learning used attributes~\cite{Farhadi09,Lampert09,jayaraman2014zero} to represent categories as vector indicating presence/absence of attributes. This vector representation can then be mapped to learn visual classifiers. Instead of using manually defined attribute-class relationships, Rohrbach et al.~\cite{Rohrbach10,Rohrbach13} mined these associations from different internet sources. Akata et al.~\cite{Akata13} used attributes as side-information to learn a semantic embedding which helps in zero-shot recognition. Recently, there have been approaches such as \cite{Qiao16} which trys to match Wikipedia text to images by modeling noise in the text description.

With the advancement of deep learning, most recent approaches can be mapped into two main research directions. The first approach is to use semantic embeddings (implicit representations). The core idea is to represent each category with learned vector representations that can be mapped to visual classifiers~\cite{Weston10,Socher13,devise12,Bernardino15,Lampert14,Fu15,Sigal16,Huang16,Kodirov17,Changpinyo16,Changpinyo17,zhang2016zero,zhang2015zero}. Socher et al.~\cite{Socher13} proposed training two different neural networks for image and language in an unsupervised manner, and then learning a linear mapping  between image representations and word embeddings. Motivated by this work, Frome et al.~\cite{devise12} proposed a system called DeViSE to train a mapping from image to word embeddings using a ConvNet and a transformation layer. By using the predicted embedding to perform nearest neighbor search, DeViSE scales up the zero-shot recognition to thousands of classes. Instead of training a ConvNet to predict the word embedding directly, Norouzi et al.~\cite{conse_iclr14} proposed another system named ConSE which constructs the image embedding by combining an existing image classification ConvNet and word embedding model. Recently, Changpinyo et al~\cite{Changpinyo16} proposed an approach to align semantic and visual manifolds via use of `phantom' classes. They report state-of-the-art results on ImageNet dataset using this approach. One strong shortcoming of these approaches is they do not use any explicit relationships between classes but rather use semantic-embeddings to represent relationships.

The second popular way to distill the knowledge is to use knowledge graph (explicit knowledge representations). Researchers have proposed several approaches on how to use knowledge graphs for object recognition~\cite{Fergus10,Salakhutdinov11,Mensink12,Palatucci09,Rohrbach11,DengECCV14,neil13,Marino17,Wu16VQA,Wang16VQA,Luijcai16}. For example, Salakhutdinov et al.~\cite{Salakhutdinov11} used WordNet  to share the representations among different object classifiers so that objects with few training examples can borrow statistical strength from related objects. On the other hand, the knowledge graph can also be used to model the mutual exclusion among different classes. Deng et al.~\cite{DengECCV14} applied these exclusion rules as a constraint in the  loss for training object classifiers (e.g. an object will not be a dog and a cat at the same time). They have also shown zero-shot applications by adding object-attribute relations into the graph. In contrast to these methods of using graph as constraints, our approach used the graph to directly generate novel object classifiers ~\cite{misra2017composing,Elhoseiny13,Ba15}.

In our work, we propose to distill information both via semantic embeddings and knowledge graphs. Specifically, given a word embedding of an unseen category and the knowledge graph that encodes explicit relationships, our approach predicts the visual classifiers of unseen categories. To model the knowledge graph, our work builds upon the Graph Convolutional Networks~\cite{GCN17}. It was originally proposed for semi-supervised learning in language processing. We extend it to our zero-short learning problem by changing the model architecture and training loss.

\vspace{-0.05in}
\section{Approach}
\vspace{-0.05in}

\begin{figure*}
    \centering
    \includegraphics[width=0.9\linewidth]{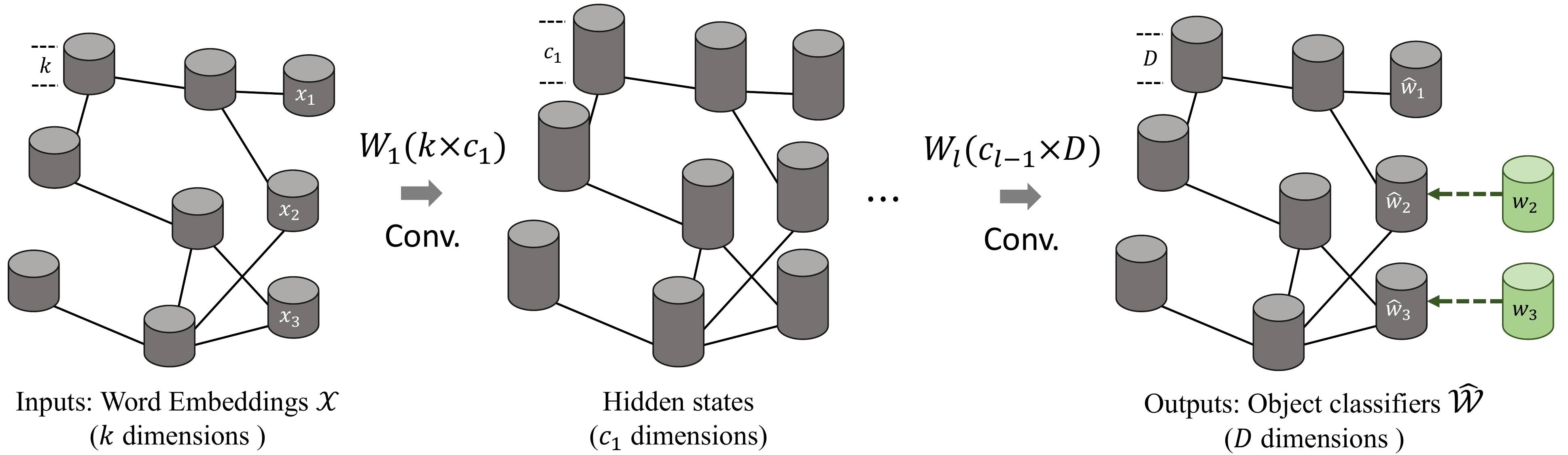}
    \caption{An example of our Graph Convolutional Network. It takes word embeddings as inputs and outputs the object classifiers. The supervision comes from the ground-truth classifiers $w_2$ and $w_3$ highlighted by green. During testing, we input the same word embeddings and obtain classifier for $x_1$ as $\hat{w}_1$. This classifier will be multiplied with the image features to produce classification scores.}\label{fig:graph}
\end{figure*}

Our goal is to distill information from both implicit (word-embeddings) and explicit (knowledge-graph)  representations for zero-shot recognition. But what is the right way to extract information? We build upon the recent work on Graph Convolutional Network (GCN)~\cite{GCN17} to learn visual classifiers. In the following, we will first introduce how the GCN is applied in natural language processing for classification tasks, and then we will go into details about our approach: applying the GCN with a regression loss for zero-shot learning.

\subsection{Preliminaries: Graph Convolutional Network} 

Graph Convolutional Network (GCN) was introduced in~\cite{GCN17} to perform semi-supervised entity classification. Given object entities, represented by word embeddings or text features, the task is to perform classification. For example,  entities such as ``dog'' and ``cat'' will be labeled as ``mammal''; ``chair'' and ``couch'' will be labeled ``furniture''. We also assume that there is a graph where nodes are entities and the edges represent relationships between entities.

Formally, given a dataset with $n$ entities $(X,Y) = \{(x_i, y_i)\}^{n}_{i=1}$ where $x_i$ represents the word embedding for entity $i$ and $y_i \in \{1,...,C\}$ represents its label. In semi-supervised setting, we know the ground-truth labels for the first $m$ entities. Our goal is to infer $y_i$ for the remaining $n-m$ entities, which do not have labels, using the word embedding and the relationship graph. In the relationship graph, each node is an entity and two nodes are linked if they have a relationship in between.

We use a function $F(\cdot)$ to represent the Graph Convolutional Network. It takes all the entity word embeddings $X$ as inputs at one time and outputs the SoftMax classification results for all of them as $F(X)$. For simplicity, we denote the output for the $i$th entity as $F_i(X)$, which is a $C$ dimension SoftMax probability vector. In training time, we apply the SoftMax loss on the first $m$ entities, which have labels as 
\begin{equation}\label{eq:softmax}
\frac{1}{m} \sum_{i=1}^{m} L_{\text{softmax}} (F_i(X), y_i).
\end{equation}
The weights of $F(\cdot)$ are trained via back-propagation with this loss. During testing time, we use the learned weights to obtain the labels for the $n-m$ entities with $F_i(X), i \in \{m+1,...,n\}$. 

Unlike standard convolutions that operate on local region in an image, in GCN the convolutional operations compute the response at a node based on the neighboring nodes defined by the adjacency graph. Mathematically, the convolutional operations for each layer in the network $F(\cdot)$ is represented  as
\begin{equation}\label{eq:conv}
Z = \hat{A} X^{\prime} W
\end{equation}
where $\hat{A}$ is a normalized version of the binary adjacency matrix $A$ of the graph, with $n \times n$ dimensions. $X^{\prime}$ is the input $n \times k$ feature matrix from the former layer. $W$ is the weight matrix of the layer with dimension $k \times c$, where $c$ is the output channel number. Therefore, the input to a convolutional layer is $n \times k$ ,and the output is a $n \times c$ matrix $Z$. These convolution operations can be stacked one after another. A  non-linear operation (ReLU) is also applied after each convolutional layer before  the features are forwarded to the next layer. For the final convolutional layer, the number of output channels is the number of label classes ($c = C$). For more details, please refer to~\cite{GCN17}. 

\subsection{GCN for Zero-shot Learning} 
Our model builds upon the Graph Convolutional Network. However, instead of entity classification, we apply it to the zero-shot recognition with a regression loss. The input of our framework is the set of categories and their corresponding semantic-embedding vectors (represented by $\mathcal{X}= \{x_i\}_{i=1}^{n}$). For the output, we want to predict the visual classifier for each input category (represented by $\mathcal{W}=\{w_i\}_{i=1}^{n}$). 

Specifically, the visual classifier we want the GCN to predict is a logistic regression model on the fixed pre-trained ConvNet features. If the dimensionality of visual-feature vector is $D$, each classifier $w_i$ for category $i$ is also a $D$-dimensional vector. Thus the output of each node in the GCN is $D$ dimensions, instead of $C$ dimensions. In the zero-shot setting, we assume that the first $m$ categories in the total $n$ classes have enough visual examples to estimate their weight vectors. For the remaining $n-m$ categories, we want to estimate their corresponding weight vectors given their embedding vectors as inputs. 

One way is to train a neural network (multi-layer perceptron) which takes $x_i$ as an input and learns to predict $w_i$ as an output. The parameters of the network can be estimated using $m$ training pairs. However, generally $m$ is small (in the order of a few hundreds) and therefore, we want to use the explicit structure of the visual world or the relationships between categories to constrain the problem. We represent these relationships as the knowledge-graph (KG).  Each node in the KG represents a semantic category.  Since we have a total of $n$ categories, there are $n$ nodes in the graph. Two nodes are linked to each other if there is a relationship between them. The graph structure is represented by the $n\times n$ adjacency matrix, $A$. Instead of building a bipartite graph as~\cite{GCN17,YangICML2016}, we replace all directed edges in the KG by undirected edges, which leads to a symmetric adjacency matrix.

As Fig.~\ref{fig:graph} shows, we use a 6-layer GCN where each layer $l$ takes as input the feature representation from previous layer ($Z_{l-1}$) and outputs a new feature representation ($Z_{l}$). For the first layer the input is $\mathcal{X}$ which is an $n \times k$ matrix ($k$ is the dimensionality of the word-embedding vector). For the final-layer the output feature-vector is $\hat{\mathcal{W}}$ which has the size of $n\times D$; $D$ being the dimensionality of the classifier or visual feature vector.

\noindent {\bf Loss-function:} For the first $m$ categories, we have predicted classifier weights $\hat{\mathcal{W}}_{1\dots m}$ and ground-truth classifier weights learned from training images $\mathcal{W}_{1\dots m}$. We use the mean-square error as the loss function between the predicted and the ground truth classifiers. 

\begin{equation}\label{eq:mse}
\frac{1}{m} \sum_{i=1}^{m} L_{\text{mse}} (\hat{w}_i, w_i).
\end{equation}

During training, we use the loss from the $m$ seen categories to estimate the  parameters for the GCN. Using the estimated parameters, we obtain the classifier weights for the zero-shot categories. At test time, we first extract the image feature representations via the pre-trained ConvNet and use these generated classifiers to perform classification on the extracted features.

\subsection{Implementation Details}
Our GCN is composed of 6 convolutional layers with output channel numbers as $2048 \rightarrow 2048 \rightarrow 1024 \rightarrow 1024 \rightarrow 512 \rightarrow D$, where $D$ represents the dimension of the object classifier. Unlike the 2-layer network presented in~\cite{GCN17}, our network is much deeper. As shown in ablative studies, we find that making the network deep is essential in generating the classifier weights. For activation functions, instead of using ReLU after each convolutional layer, we apply LeakyReLU~\cite{Maas13,xu15} with the negative slope of $0.2$. Empirically, we find that LeakyReLU leads to faster convergence for our regression problem. 

While training our GCN, we perform L2-Normalization on the outputs of the networks and the ground-truth classifiers. During testing, the generated classifiers of unseen classes are also L2-Normalized. We find adding this constraint important, as it regularizes the weights of all the classifiers into similar magnitudes. In practice, we also find that the last layer classifiers of the ImageNet pre-trained networks are naturally normalized. That is, if we perform L2-Normalization on each of the last layer classifiers during testing, the performance on the ImageNet 2012 1K-class validation set changes marginally ($<1\%$).

To obtain the word embeddings for GCN inputs, we use the GloVe text model~\cite{pennington2014glove} trained on the Wikipedia dataset, which leads to 300-d vectors. 
For the classes whose names contain multiple words, we match all the words in the trained model and find their embeddings. By averaging these word embeddings, we obtain the class embedding.

\section{Experiment}
\vspace{-0.05in}
We now perform experiments to showcase that our approach: (a) improves the state-of-the-art by a significant margin; (b) is robust to different pre-trained  ConvNets and noise in the KG. We use two datasets in our experiments. The first dataset we use is constructed from publicly-available knowledge bases. The dataset consists of relationships and graph from  Never-Ending Language Learning (NELL)~\cite{carlson-aaai} and images from Never-Ending Image Learning (NEIL)~\cite{neil13}. This is an ideal dataset for: (a) demonstrating that our approach is robust even with automatically learned (and noisy) KG; (b) ablative studies since the KG in this domain is rich, and we can perform ablations on KG as well.

Our final experiments are shown on the standard ImageNet dataset. We use the same settings as the baseline approaches~\cite{devise12,conse_iclr14,Changpinyo16} together with the WordNet~\cite{miller1995wordnet} knowledge graph. We show that our approach surpasses the state-of-the-art methods by a significant margin. 

\subsection{Experiments on NELL and NEIL} 
\vspace{-0.05in}
\noindent
\textbf{Dataset settings.} For this experiment, we construct a new knowledge graph based on the NELL~\cite{carlson-aaai} and NEIL~\cite{neil13} datasets. Specifically, the object nodes in NEIL correspond to the nodes in NELL. The NEIL dataset offers the sources of images and the NELL dataset offers the common sense knowledge rules. However, the NELL graph is incredibly large \footnote{http://rtw.ml.cmu.edu/}: it contains roughly 1.7M types of object entities and around 2.4M edges representing the relationships between every two objects. Furthermore, since NELL is constructed automatically, there are noisy edges in the graph. Therefore, we create sub-graphs for our experiments. 

The process of constructing this sub-graph is straightforward. We perform Breadth-first search (BFS) starting from the NEIL nodes. We discover paths with maximum length $K$ hops such that the first and last node in the path are NEIL nodes. We add all the nodes and edges in these paths into our  sub-graph. We set $K=7$ during BFS because we discover a path longer than 7 hops will cause the connection between two objects noisy and unreasonable. For example, ``jeep'' can be connected to ``deer'' in a long path but they are hardly semantically related.

Note that each edge in NELL has a confidence value that is usually larger than $0.9$. For our experiments, we create two different versions of sub-graphs. The first smaller version is a graph with high value edges (larger than $0.999$), and the second one used all the edges regardless of their confidence values. The statistics of the two sub-graphs are summarized in Table~\ref{tab:dataset}. For the larger sub-graph, we have 14K object nodes. Among these nodes, 704 of them have corresponding images in the NEIL database. We use 616 classes for training our GCN and leave 88 classes for testing. Note that these 88 testing classes are randomly selected among the classes that have no overlap with the 1000 classes in the standard ImageNet classification dataset. The smaller knowledge graph is around half the size of the larger one. We use the same 88 testing classes in both settings

\begin{table}[t]
\begin{center}
\begin{tabular}{l|c|c|c}
\hline
   ~            & All                   & NEIL Nodes      &  ~          \\
Dataset &  Nodes & (Train/Test)  & Edges  \\
\hline
High Value Edges    & 8819                        & 431/88                     &  40810           \\
All Edges        & 14612                       & 616/88                     &  96772           \\
\hline

\end{tabular}
\end{center}
\vspace{-0.05in}
\caption{Dataset Statistics: Two different sizes of knowledge graphs in our experiment.}
\label{tab:dataset}
\vspace{-0.2in}
\end{table}

\noindent
\textbf{Training details.} For training the ConvNet on NEIL images, we use the 310K images associated with the 616 training classes. The evaluation is performed on the randomly selected 12K images associated with the 88 testing classes, i.e. all images from the training classes are excluded during testing. We fine-tune the ImageNet pre-trained VGGM~\cite{Chatfield14} network architecture with relatively small $fc7$ outputs (128-dimension). Thus the object classifier dimension in $fc8$ is $128$. For training our GCN, we use the ADAM~\cite{Kingma14adam} optimizer with learning rate $0.001$ and weight decay $0.0005$. We train our GCN for 300 epochs for every experiment.

\noindent
\textbf{Baseline method.} We compare our method with one of the state-of-the-art methods, ConSE~\cite{conse_iclr14}, which shows slightly better performance than DeViSE~\cite{devise12} in ImageNet. As a brief introduction, ConSE first feedforwards the test image into a ConvNet that is trained only on the training classes. With the output probabilities, ConSE selects top $T$ predictions $\{p_i\}_{i=1}^{T}$ and the word embeddings $\{x_i\}_{i=1}^{T}$~\cite{word2vec} of these classes. It then generates a new word embedding by weighted averaging the $T$ embeddings with the probability $\frac{1}{T} \sum_{i = 1}^{T} p_i x_i $. This new embedding is applied to perform nearest neighbors in the word embeddings of the testing classes. The top retrieved classes are selected as the final result. We enumerate different values of $T$ for evaluations.

\begin{table}[t]
\begin{center}
\begin{tabular}{l|l|ccccc}
\hline
~                                                          & ~                    & \multicolumn{4}{c}{ Hit@$k$ (\%)} \\
\cline{3-6}
Test Set                                                  & Model                & 1     & 2     & 5     & 10      \\
\hline
\multirow{4}{*}{High Value }                  & \softmaxpp{(5)}          & 6.6 &   9.6 &   13.6 &  19.4 \\
\multirow{4}{*}{Edges}                        & \softmaxpp{(10)}         & 7.0 &   9.8 &   14.2 & 20.1  \\
~                                             & \softmaxpp{(431)}        & 6.7 &   9.7 &   14.9 &   20.5 \\
~                                             & Ours             & \bf9.1 & \bf	16.8 & \bf	23.2	 & \bf	47.9 \\
\hline
\multirow{4}{*}{All Edges}                 & \softmaxpp{(5)}          & 7.7 &   10.1 &   13.9 &   19.5 \\
~                                          & \softmaxpp{(10)}         & 7.7 &   10.4 &   14.7 &   20.5  \\
~                                          & \softmaxpp{(616)}        & 7.7 &   10.5 &   15.7 &   21.4 \\
~                                          & Ours                   &\bf10.8	&\bf18.4	&\bf33.7 &\bf	49.0\\

\hline

\end{tabular}
\end{center}
\vspace{-0.05in}
\caption{Top-k accuracy for different models in different settings.}
\label{tab:zeroshot-hit}
\end{table}

\noindent
\textbf{Quantitative Results.} We perform evaluations on the task of 88 unseen categories classification. Our metric is based on the percentage of correctly retrieved test data (out of top $k$ retrievals) for a given zero-shot class. The results are shown in Table~\ref{tab:zeroshot-hit}. We evaluate our method on two different sizes of knowledge graphs. We use ``High Value Edges'' to denote the knowledge graph constructed based on high confidence edges. ``All Edges'' represents the graph constructed with all the edges. We denote the baseline~\cite{conse_iclr14} as ``ConSE(T)'' where we set $T$ to be 5, 10 and the number of training classes.

Our method outperforms the ConSE baseline by a large margin. In the ``All Edges'' dataset, our method outperforms ConSE $3.6\%$ in top-1 accuracy. {\bf More impressively, the accuracy of our method is almost 2 times as that of ConSE in top-2 metric and even more than 2 times in top-5 and top-10 accuracies}. These results show that using knowledge graph with word embeddings in our method leads to much better result than the state-of-the-art results with word embeddings only.

\noindent
\textbf{From small to larger graph.} In addition to improving performance in zero-shot recognition, our method obtains more performance gain as our graph size increases. As shown in Table~\ref{tab:zeroshot-hit},  our method performs better by switching from the small to larger graph. Our approach has obtained $2 \sim 3\%$ improvements in all the metrics. On the other hand, there is little to no improvements in ConSE performance. It also shows that the KG does not need to be hand-crafted or cleaned. Our approach is able to robustly handle the errors in the graph structure.

\begin{figure}[t]
    \centering
    \includegraphics[width=0.95\linewidth]{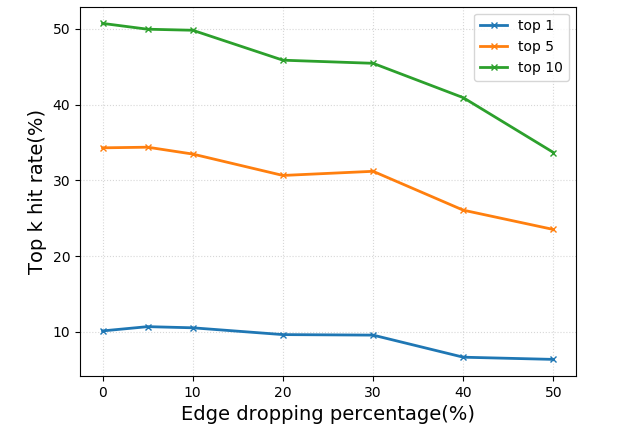}
    \caption{We randomly drop  $5\%$ to $50\%$ of the edges in the ``All Edges'' graph and show the top-1, top-5 and top-10 accuracies. }\label{fig:drop}
    \vspace{-0.1in}
\end{figure}

\noindent
\textbf{Resilience to Missing Edges} We explore how the performance of our model changes if we randomly drop $5\%$ to $50\%$ of the edges in the ``All Edges'' graph. As Fig.~\ref{fig:drop} shows, by dropping from $5\%$ to $10\%$ of edges, the performance of our model changes negligibly. This is mainly because the knowledge graph can have redundant information with 14K nodes and 97K edges connecting them. This again implies that our model is robust to small noisy changes in the graph. As we start deleting more than $30\%$ of the edges, the accuracies drop drastically. This indicates that the performance of our model is highly correlated to the size of the knowledge graph. 

\noindent
\textbf{Random Graph?} It is clear that our approach can handle noise in the graph. But does any random graph work? To demonstrate that the structure of the graph is still critical we also created some trivial graphs: (i) star model: we create a graph with one single root node and only have edges connecting object nodes to the root node; (ii) random graph: all nodes in the graph are randomly connected. Table~\ref{tab:trivial-kg} shows the results. It is clear that all the numbers are close to random guessing, which means a reasonable graph plays an important role and a random graph can have negative effects on the model. 

\begin{table}[h]
\begin{center}
\begin{tabular}{l|l|ccccc}
\hline
~                                                         & ~                    & \multicolumn{4}{c}{ Hit@$k$ (\%)} \\
\cline{3-6}
Test Set                                                  & Trivial KG                 & 1     & 2     & 5     & 10      \\
\hline
\multirow{3}{*}{All Edges} 
                        & Star Model        & 1.1 &    1.6 &    4.8 &   9.7 \\
~                       & Random Graph        & 1.0 &    2.2 &   5.6 &   11.3 \\
\hline

\end{tabular}
\end{center}
\vspace{-1ex}
\caption{Top-k accuracy on trivial knowledge graphs we create.}
\label{tab:trivial-kg}
\vspace{-1ex}
\end{table}

\begin{figure}[t]
    \centering
    \includegraphics[width=1.01\linewidth]{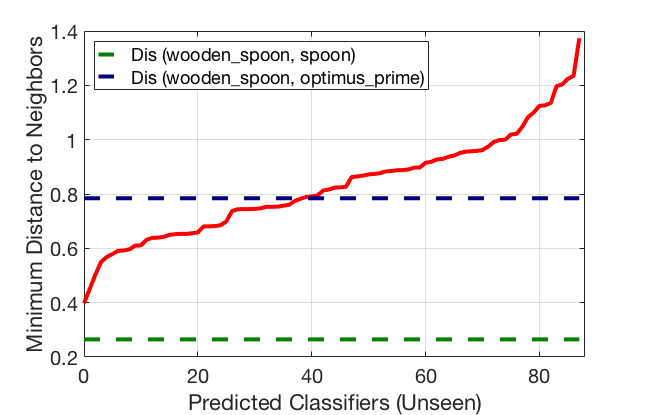}
    \caption{We compute the minimum Euclidean distances between predicted and training classifiers. The distances are plotted by sorting them from small to large. }\label{fig:ablative}
    \vspace{-0.1in}
\end{figure}

\begin{figure}[b]
    \centering
    \vspace{-0.15in}
    \includegraphics[width=0.9\linewidth]{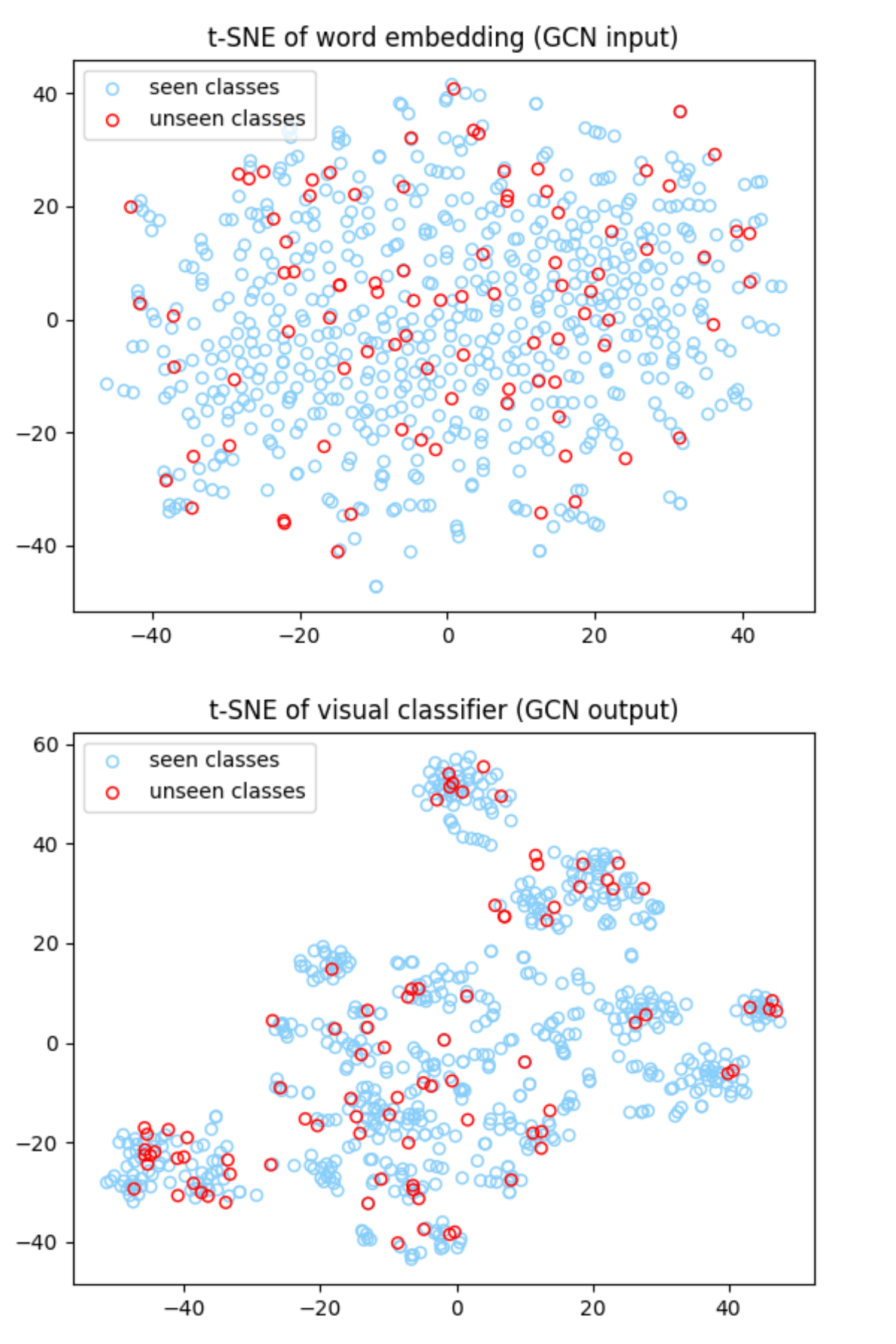}
    \caption{t-SNE visualizations for our word embeddings and GCN output visual classifiers in the ``All Edges'' dataset. The test classes are shown in red.}\label{fig:tsne}
\end{figure}

\noindent
\textbf{How important is the depth of GCN?} We show that making the Graph Convolutional Network deep is critical in our problem. We show the performance of using different numbers of layers for our model on the ``All Edges'' knowledge graph shown in Table~\ref{tab:ablative-hit}. For the 2-layer model we use $512$ hidden neurons, and the 4-layer model has output channel numbers as $2048 \rightarrow 1024 \rightarrow 512 \rightarrow 128$. We show that the performance keeps increasing as we make the model deeper from 2-layer to 6-layer. The reason is that increasing the times of convolutions is  essentially increasing the  times of message passing between nodes in the graph. However, we do not observe much gain by adding more layers above the 6-layer model. One potential reason might be that the optimization becomes harder as the network goes deeper.  

\begin{table}[t]
\begin{center}
\begin{tabular}{l|l|ccccc}
\hline
~                                                         & ~                    & \multicolumn{4}{c}{ Hit@$k$ (\%)} \\
\cline{3-6}
Test Set                                                  & Model                & 1     & 2     & 5     & 10      \\
\hline
\multirow{3}{*}{All Edges}  
                        & Ours (2-layer)        & 5.3 &    8.7 &    15.5 &   24.3 \\
~                       & Ours (4-layer)        & 8.2 &    13.5 &   27.1 &  41.8  \\
~                       & Ours (6-layer)    &\bf10.8	&\bf18.4	&\bf33.7 &\bf	49.0\\  
\hline

\end{tabular}
\end{center}
\vspace{-0.05in}
\caption{Top-k accuracy with different depths of our model.}
\label{tab:ablative-hit}
\vspace{-0.2in}
\end{table}

\begin{table*}[t]
\begin{center}

\subfloat[Top-k accuracy for different models when testing on only unseen classes. \label{tab:ablation:testontest}]{
\tablestyle{3.4pt}{1.00}
\begin{tabular}{c|l|l|cccccc}
\hline
~       & ~     & ~         & \multicolumn{5}{c}{ Hit@$k$ (\%)} \\
\cline{4-8}
Test Set                                                  & Model  & ConvNets   & 1     & 2     & 5     & 10   & 20      \\
\hline

\multirow{4}{*}{2-hops }                       & ConSE~\cite{Changpinyo16}  & Inception-v1      & 8.3  &   12.9 &   21.8 &  30.9 & 41.7 \\
~                                             & ConSE(us) & Inception-v1   & 12.4 &   18.4 &   25.3 &  28.5 & 31.8 \\
~                                             & SYNC~\cite{Changpinyo16}  & Inception-v1       & 10.5 &   17.7 &   28.6 &  40.1 & 52.0 \\
~                                             & EXEM~\cite{Changpinyo17}  & Inception-v1       & 12.5 &   19.5 &   32.3 &  43.7 & 55.2 \\
\cline{2-8}
~                                             & Ours  & Inception-v1       & 18.5 & 31.3 & 50.1 & 62.4 & 72.0  \\
~                                             & Ours  & ResNet-50       & \bf 19.8 & \bf 33.3 & \bf 53.2 & \bf 65.4 & \bf 74.6  \\
\hline
\multirow{4}{*}{3-hops }                       & ConSE~\cite{Changpinyo16}  & Inception-v1      & 2.6  &   4.1 &   7.3 &  11.1 & 16.4 \\
~                                             & ConSE(us) & Inception-v1   & 3.2  &   4.9 &   7.6 &  9.7  & 11.4 \\
~                                             & SYNC~\cite{Changpinyo16}  & Inception-v1       & 2.9  &   4.9 &   9.2 &  14.2 & 20.9 \\
~                                             & EXEM~\cite{Changpinyo17}  & Inception-v1       & 3.6  &   5.9 &   10.7 & 16.1 & 23.1 \\
\cline{2-8}
~                                             & Ours  & Inception-v1       & 3.8  &   6.9 &   13.1 & 18.8 & 26.0  \\
~                                             & Ours  & ResNet-50       & \bf 4.1 & \bf 7.5 & \bf 14.2 & \bf 20.2 & \bf 27.7  \\
\hline
\multirow{4}{*}{All }                         & ConSE~\cite{Changpinyo16}  & Inception-v1      & 1.3  &   2.1 &   3.8 &  5.8 & 8.7 \\
~                                             & ConSE(us) & Inception-v1                     & 1.5	&2.2	&3.6	&4.6	&5.7 \\
~                                             & SYNC~\cite{Changpinyo16}  & Inception-v1       & 1.4 &   2.4 &   4.5 &  7.1 & 10.9 \\
~                                             & EXEM~\cite{Changpinyo17}  & Inception-v1       & \bf 1.8 &   2.9 &   5.3 &  8.2 & 12.2 \\
\cline{2-8}
~                                             & Ours  & Inception-v1       & 1.7 & 3.0 & 5.8 & 8.4 & 11.8  \\
~                                             & Ours  & ResNet-50       & \bf 1.8 & \bf 3.3 & \bf 6.3 & \bf 9.1 & \bf 12.7  \\
\hline
\end{tabular} }\hspace{3mm}
\subfloat[Top-k accuracy for different models when testing on both seen and unseen classes (a more practical and generalized setting). \label{tab:ablation:testonboth}]{
\tablestyle{3.4pt}{1.00}
\begin{tabular}{c|l|l|cccccc}
\hline
~       & ~     & ~         & \multicolumn{5}{c}{ Hit@$k$ (\%)} \\
\cline{4-8}
Test Set                                                  & Model  & ConvNets   & 1     & 2     & 5     & 10   & 20      \\
\hline

\multirow{4}{*}{2-hops }                       & DeViSE~\cite{devise12}  & AlexNet      & 0.8  &   2.7 &   7.9 &  14.2 & 22.7 \\
\multirow{4}{*}{(+1K) }                                             & ConSE~\cite{conse_iclr14} & AlexNet  & 0.3 &   6.2 &   17.0 &  24.9 & 33.5 \\
~                                             & ConSE(us)  & Inception-v1       & 0.2 &   7.8 &   18.1 &  22.8 & 26.4 \\
~                                             & ConSE(us)  & ResNet-50       & 0.1 &  11.2 &   24.3 &  29.1  & 32.7 \\
\cline{2-8}
~                                             & Ours  & Inception-v1      &7.9	&18.6		&39.4	&53.8	&65.3  \\
~                                             & Ours  & ResNet-50       & \bf 9.7 & \bf 20.4 & \bf 42.6 & \bf 57.0 & \bf 68.2  \\
\hline
\multirow{4}{*}{3-hops}                       & DeViSE~\cite{devise12}  & AlexNet      & 0.5  &   1.4 &   3.4 &  5.9 & 9.7 \\
\multirow{4}{*}{(+1K) }                                            & ConSE~\cite{conse_iclr14} & AlexNet  & 0.2 &   2.2 &   5.9 &  9.7 & 14.3 \\
~                                             & ConSE(us)  & Inception-v1       & 0.2 &   2.8 &   6.5 &  8.9 & 10.9 \\
~                                             & ConSE(us)  & ResNet-50       & 0.2 &   3.2 &   7.3 &  10.0 & 12.2 \\
\cline{2-8}
~                                             & Ours  & Inception-v1       & 1.9 & 4.6 & 10.9 & 16.7 & 24.0  \\
~                                             & Ours  & ResNet-50       & \bf 2.2 & \bf 5.1 & \bf 11.9 & \bf 18.0 & \bf 25.6  \\
\hline
\multirow{4}{*}{All}                         & DeViSE~\cite{devise12}  & AlexNet      & 0.3  & 0.8 & 1.9 & 3.2 & 5.3 \\
\multirow{4}{*}{(+1K) }                       & ConSE~\cite{conse_iclr14} & AlexNet        & 0.2  & 1.2 & 3.0 & 5.0 & 7.5 \\
~                                             & ConSE(us)  & Inception-v1      &0.1	&1.3	&3.1	&4.3	&5.5 \\
~                                             & ConSE(us)  & ResNet-50       &0.1	&1.5	&3.5	&4.9	&6.2 \\
\cline{2-8}
~                                             & Ours  & Inception-v1       & 0.9 & 2.0 & 4.8 & 7.5 & 10.8  \\
~                                             & Ours  & ResNet-50       & \bf 1.0 & \bf 2.3 & \bf 5.3 & \bf 8.1 & \bf 11.7  \\
\hline
\end{tabular} }\hspace{3mm}
\caption{Results on ImageNet. We test our model on 2 different settings over 3 different datasets.}
\vspace{-0.2in}
\end{center}
\end{table*}

\noindent
\textbf{Is our network just copying classifiers as outputs?} Even though we show our method is better than ConSE baseline, is it possible that it learns to selectively copy the nearby classifiers? To show our method is not learning  this trivial solution, we compute the Euclidean distance between our generated classifiers and the training classifiers. More specifically, for a generated classifier, we compare it with the classifiers from the training classes that are at most 3-hops away. We calculate the minimum distance between each generated classifier and its neighbors. We sort the distances for all $88$ classifiers and plot  Fig.~\ref{fig:ablative}. As for reference, the distance between ``wooden\_spoon'' and ``spoon'' classifiers in the training set is $0.26$  and the distance between ``wooden\_spoon'' and ``optimus\_prime'' is $0.78$. We can see that our predicted classifiers are quite different from its neighbors.

\noindent
\textbf{Are the outputs only relying on the word embeddings?}  We perform t-SNE~\cite{maaten2008visualizing} visualizations to show that our output classifiers are not just derived from the word embeddings. We show the t-SNE~\cite{maaten2008visualizing} plots of both the word embeddings and the classifiers of the seen and unseen classes in the ``All Edges'' dataset. As Fig.~\ref{fig:tsne} shows, we have very different clustering results between the word embeddings and the object classifiers, which indicates that our GCN is not just learning a direct projection from word embeddings to classifiers.

\subsection{Experiments on WordNet and ImageNet} 
\vspace{-0.05in}
We now perform our experiments on a much larger-scale ImageNet~\cite{russakovsky2015imagenet} dataset. We adopt the same train/test split settings as~\cite{devise12,conse_iclr14}. More specifically, we report our results on 3 different test datasets: ``2-hops'', ``3-hops'' and the whole ``All'' ImageNet set. These datasets are constructed according to  how similar the classes are related to the classes in the ImageNet 2012 1K dataset. For example, ``2-hops'' dataset (around 1.5K classes) includes the classes from the ImageNet 2011 21K set which are semantically very similar to the ImageNet 2012 1K classes. ``3-hops'' dataset (around 7.8K classes) includes the classes that are within 3 hops of the ImageNet 2012 1K classes, and the ``All'' dataset includes all the labels in ImageNet 2011 21K. There are no common labels between the ImageNet 1K class and the classes in these 3-dataset. It is also obvious to see that as the number of class increases, the task becomes more challenging. 

As for knowledge graph, we use the sub-graph of the WordNet~\cite{miller1995wordnet}, which includes around 30K object nodes\footnote{http://www.image-net.org/explore}. 

\noindent
\textbf{Training details.} Note that to perform testing on 3 different test sets, we only need to train one set of ConvNet and GCN. We use two different types of ConvNets as the base network for computing visual features: Inception-v1~\cite{szegedy2015going} and ResNet-50~\cite{resnet}. Both networks are pre-trained using the ImageNet 2012 1K dataset and no fine-tuning is required. For Inception-v1, the output feature of the second to the last layer has $1024$ dimensions, which leads to $D=1024$ object classifiers in the last layer. For ResNet-50, we have $D=2048$. Except for the changes of output targets, other settings of training GCN remain the same as those of the previous experiments on NELL and NEIL. It is worthy to note that our GCN model is robust to different sizes of outputs. The model shows consistently better results as the representation (features) improves from Inception-v1 ($68.7\%$ top-1 accuracy in ImageNet 1K val set) to ResNet-50 ($75.3\%$). 

We evaluate our method with the same metric as the previous experiments: the percentage of hitting the ground-truth labels among the top $k$ predictions. However, instead of only testing with the unseen object classifiers, we include both training and the predicted classifiers during testing, as suggested by~\cite{devise12,conse_iclr14}. Note that in these two settings of experiments, we still perform testing on the same set of images associated with unseen classes only.

\begin{table}[t]
\begin{center}{
\begin{tabular}{l|l|ccccc}
\hline
 ~     & Word    & \multicolumn{5}{c}{ Hit@$k$ (\%)} \\
\cline{3-7}
Model    & Embedding & 1     & 2     & 5     & 10   & 20      \\
\hline

\cite{zhang2017learning} & GloVe  & 7.8  & 11.5 & 17.2 & 21.2 & 25.6 \\
Ours          	             & GloVe  & \bf 18.5 & \bf 31.3 & \bf 50.1 & \bf 62.4 & \bf 72.0 \\
\hline
\cite{zhang2017learning}  & FastText & 9.8 & 16.4 & 27.8 & 37.6 & 48.4 \\
Ours                          & FastText & \bf 18.7 & \bf 30.8 & \bf 49.6 & \bf 62.0 & \bf 71.5 \\
\hline
\cite{zhang2017learning}  & GoogleNews & 13.0 & 20.6 & 33.5 & 44.1 & 55.2 \\
Ours 		 	              & GoogleNews &\bf 18.3 & \bf 31.6 & \bf 51.1 & \bf 63.4 & \bf 73.0 \\
\hline
\end{tabular} }
\caption{Results with different word embeddings on ImageNet (2 hops), corresponding to the experiments in Table~\ref{tab:ablation:testontest}. }
\vspace{-0.1in}
\end{center}
\label{tab:differentword}
\vspace{-0.2in}
\end{table}

\noindent
\textbf{Testing without considering the training labels.} We first perform experiments excluding the classifiers belonging to the training classes during testing. We report our results in Table.~\ref{tab:ablation:testontest}. We  compare our results to the  recent state-of-the-art methods SYNC~\cite{Changpinyo16} and EXEM~\cite{Changpinyo17}. We show experiments with the same pre-trained ConvNets (Inception-v1) as~\cite{Changpinyo16,Changpinyo17}.
Due to unavailability of their word embeddings for all the nodes in KG, we use a different set of word embeddings (GloVe) ,which is publicly available. 

Therefore, we first investigate if the change of word-embedding is crucial. We show this via the ConSE baseline. Our re-implementation of ConSE, shown as ``ConSE(us)'' in the table, uses the GloVe whereas the ConSE method implemented in~\cite{Changpinyo16,Changpinyo17} uses their own word embedding. We see that both approaches have similar performance. Ours is slightly better in top-1 accuracy while the one in~\cite{Changpinyo16,Changpinyo17} is better in top-20 accuracy. Thus, with respect to zero-shot learning, both word-embeddings seem equally powerful.

We then compare our results with SYNC~\cite{Changpinyo16} and EXEM~\cite{Changpinyo17}. With the same pre-trained ConvNet Inception-v1, our method outperforms almost all the other methods on all the datasets and metrics. On the ``2-hops'' dataset,  our approach outperforms all methods with a large margin: around $6\%$ on top-1 accuracy and $17\%$ on top-5 accuracy.  On the ``3-hops'' dataset, our approach is consistently better than EXEM~\cite{Changpinyo17} around $2 \sim 3\%$ from top-5 to top-20 metrics. 

By replacing the Inception-v1 with the ResNet-50, we obtain another performance boost in all metrics. For the top-5 metric,  our final model outperforms the state-of-the-art method EXEM~\cite{Changpinyo17} by a whooping $20.9\%$ in the ``2-hops'' dataset, $3.5\%$ in the ``3-hops'' dataset and $1\%$ in the ``All'' dataset. Note that the gain is diminishing because the task increases in difficulty as the number of unseen classes increases.

\noindent
\textbf{Sensitivity to word embeddings.} Is our method sensitive to word embeddings? What will happen if we use different word embeddings as inputs? We investigate 3 different word embeddings including GloVe~\cite{pennington2014glove} (which is used in the other experiments in the paper), FastText~\cite{joulin2016fasttext} and word2vec~\cite{word2vec} trained with GoogleNews. As for comparisons, we have also implemented the method in~\cite{zhang2017learning} which trains a direct mapping from word embeddings to visual features without knowledge graphs. We use the Inception-v1 ConvNet to extract visual features. We show the results on ImageNet (with the 2-hops setting same as Table~\ref{tab:ablation:testontest}). We can see that~\cite{zhang2017learning} highly relies on the quality of the word embeddings (top-5 results range from $17.2\%$ to $33.5\%$). On the other hand, our top-5 results are stably around $50\%$ and are much higher than~\cite{zhang2017learning}. With the GloVe word embeddings, our approach has {\bf a relative improvement of almost 200\%} over ~\cite{zhang2017learning}. This again shows graph convolutions with knowledge graphs play a significant role in improving zero-shot recognition.

\noindent
\textbf{Testing with the training classifiers.} Following the suggestions in~\cite{devise12,conse_iclr14}, a more practical setting for zero-shot recognition is to include both seen and unseen category classifiers during testing. We test our method in this generalized setting. Since there are very few baselines available for this setting of experiment, we can only compare the results with ConSE and DeViSE. We have also re-implemented the ConSE baselines with both Inception-v1 and ResNet-50 pre-trained networks. As Table~\ref{tab:ablation:testonboth} shows
our method almost doubles the performance compared to the baselines on every metric and all 3-datasets. Moreover, we can still see the boost in of performance by switching the pre-trained Inception-v1 network to ResNet-50.

\begin{figure}[t]
    \centering
    \vspace{-0.15in}
    \includegraphics[width=0.95\linewidth]{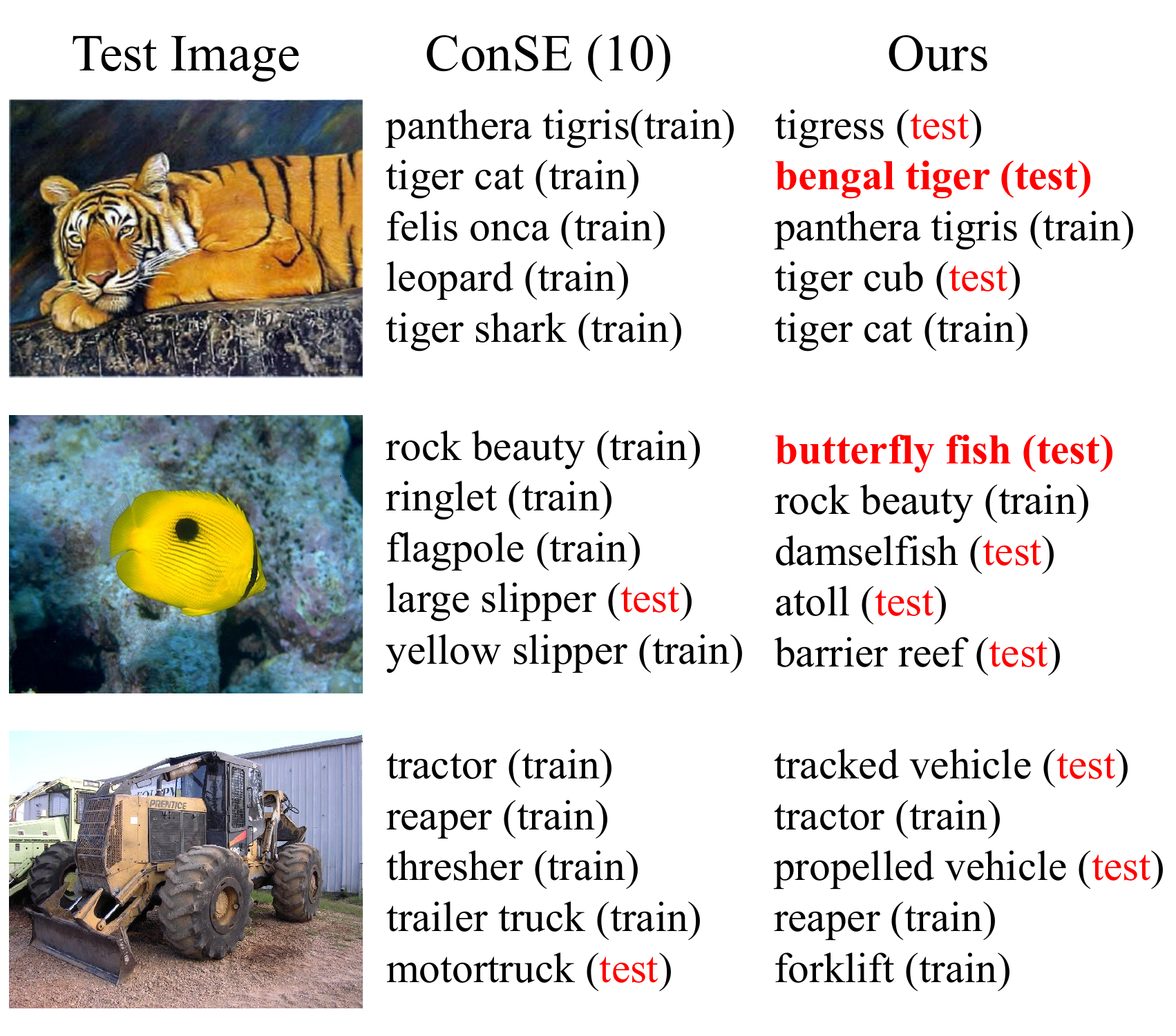}
    \caption{Visualization of top 5 prediction results for 3 different  images. The correct prediction results are highlighted by red bold characters. The unseen classes are  marked with a red ``test'' in the bracket. Previously seen classes have a plain ``train''  in the bracket. }
    \label{fig:visualization}
    \vspace{-0.15in}
\end{figure}

\noindent
\textbf{Visualizations.} We finally perform visualizations using our model and ConSE with $T=10$ in Fig. ~\ref{fig:visualization} (Top-5 prediction results). We can see that our method significantly outperforms ConSE(10) in these examples. Although ConSE(10) still gives reasonable results in most cases,  the  output labels are biased to be within the training labels. On the other hand, our method outputs the unseen classes as well.

\vspace{-0.1in}
\section{Conclusion}
\vspace{-0.1in}
We have presented an approach for zero-shot recognition using the semantic embeddings of a category and the knowledge graph that encodes the relationship of the novel category to familiar categories. Our work also shows that  a knowledge graph provides supervision to learn meaningful classifiers on top of semantic embeddings. Our results indicate a significant improvement over current state-of-the-art.

{\footnotesize
{\noindent {\bf Acknowledgement}: This work was supported by ONR MURI N000141612007, Sloan, Okawa Fellowship to AG and NVIDIA Fellowship to XW. We would also like to thank Xinlei Chen, Senthil Purushwalkam, Zhilin Yang and Abulhair Saparov for many helpful discussions.}
}

{\small
\bibliographystyle{ieee}
\bibliography{local}
}

\end{document}